# Linear Contour Learning: A Method for Supervised Dimension Reduction

Bing Li
Department of Statistics
Pennsylvania State University
University Park, PA 16802
bing@stat.psu.edu

Hongyuan Zha
Department of Computer Science
Pennsylvania State University
University Park, PA 16802
zha@cse.psu.edu

Francesca Chiaromonte
Department of Statistics
Pennsylvania State University
University Park, PA 16802
chiaro@stat.psu.edu

## Abstract

We propose a novel approach to sufficient dimension reduction in regression, based on estimating contour directions of negligible variation for the response surface. These directions span the orthogonal complement of the minimal space relevant for the regression, and can be extracted according to a measure of the variation in the response, leading to *General Contour Regression* (GCR). In comparison to existing sufficient dimension reduction techniques, this contour-based methodology guarantees exhaustive estimation of the central space under ellipticity of the predictor distribution and very mild additional assumptions, while maintaining $\sqrt{n}$-consistency and computational ease. Moreover, it proves to be robust to departures from ellipticity. We also establish some useful population properties for GCR. Simulations to compare performance with that of standard techniques such as ordinary least squares, sliced inverse regression, principal hessian directions, and sliced average variance estimation confirm the advantages anticipated by theoretical analyses. We also demonstrate the use of contour-based methods on a data set concerning grades of students from Massachusetts colleges.

## 1 Introduction and Background

Dimension reduction methods have the potential to reduce computational cost and storage requirements, and improve the performance of various learning algorithms. Within the machine learning and AI fields, most research on dimension reduction focuses on unsupervised approaches; here we consider dimension reduction for the regression of a continuous response $Y$ on a vector of continuous predictors $X = (X_1, \ldots, X_p)^T \in \mathbb{R}^p$. Our approach is based on *sufficient dimension reduction*, a body of statistical theory and methods for reducing the dimension of $X$ while preserving information on the regression; that is, on the conditional distribution of $Y|X$. A *dimension reduction subspace* (Cook, 1998) is defined as the column span of any $p \times d$ ($d \leq p$) matrix $\eta$ such that

$$Y \perp\!\!\!\perp X | \eta^T X \tag{1}$$

where $\perp\!\!\!\perp$ indicates independence, i.e., conditioning upon $\eta^T X$, $Y$ is independent of $X$. Note that the conditional independence in (1) is not affected by multiplying $\eta$ from the right with a full rank matrix; what matters in this relation is the column space, and not the specific form, of $\eta$.

A regression can admit many subspaces satisfying (1), because if it holds for $\eta$ then it also holds for any other matrix whose column space includes that of $\eta$. Naturally, we are interested in the subspace with the minimal dimension. Though special instances of regressions can be constructed admitting more than one minimal dimension reduction subspace, under mild conditions that are almost always verified in practice, the minimal subspace is uniquely defined and coincides with the intersection of all subspaces satisfying (1). See (Cook, 1998 and Chiaromone and Cook, 2002). This intersection is referred to as the central subspace, and is denoted by $\mathcal{S}_{Y|X}$. The dimension of $\mathcal{S}_{Y|X}$ is called the *structural dimension* and is denoted by $q$. Let $\beta$ be a $p \times q$ matrix whose columns span the central subspace. Then, $\mathcal{S}_{Y|X}$ can be represented by span$(\beta)$, the subspace of $\mathbb{R}^p$ spanned by the columns of $\beta$.

The central subspace, as represented by span$(\beta)$, can be estimated without estimating a response surface or introducing strong assumptions of the form of dependence between $Y$ and $X$. Well known estimation methods include Ordinary Least Squares (OLS; Li and Duan, 1989), Sliced Inverse Regression (SIR; Li, 1991), Principal Hessian Directions (PHD; Li, 1992), and Sliced Average Variance Estimation (SAVE; Cook and Weisberg, 1991). These dimension reduction methods constitute effective pre-modeling tools to replace high-dimensional regressions with informationally equivalent ones comprising only a few linear combinations of the original predictors. The reduction greatly facilitates model building, as well as the use of non-parametric techniques. Moreover, dimension reduction methods allow a comprehensive visualization of the data whenever the estimated structural dimension is 1, 2 or possibly 3, which is the case in a vast majority of practical applications. In this sense, sufficient dimension reduction provides a foundation for regression graphics, as argued in Cook (1998) and Chiaromonte and Cook (2002).

All the above-mentioned methods are $\sqrt{n}$-consistent and computationally inexpensive, due to the fact that they exploit *global* features of the dependence of $Y$ on $X$, as captured by mixed moments estimated on the data. OLS estimates one direction in the central space as a sample version of $\Sigma^{-1/2}\sigma_{X,Y}$, where $\Sigma$ is the variance matrix of $X$ and $\sigma_{X,Y}$ the covariance vector between $X$ and $Y$. SIR and SAVE use an *inverse* approach, considering mean and variance functions for the regression of the standardized predictor $Z = \Sigma^{-1/2}(X - E(X))$ on the response $Y$. In particular, SIR estimates directions through eigenvectors of a sample version of $E[E(Z|Y)E(Z|Y)^T]$, while SAVE uses eigenvectors of a sample version of $E[(I - Var(Z|Y))(I - Var(Z|Y))^T]$. In the most widespread implementations of both SIR and SAVE, sample versions of the relevant matrices are obtained with a system of slices on $Y$. Back to a forward regression approach, PHD considers the Hessian matrix $H(X)$ of the mean function $E(X|Y)$, and estimates directions in the central space through eigenvectors of a sample version of $E[H(X)]\Sigma$. Detailed descriptions of these methods and additional references are provided in Cook (1998).

Global methods also have common limitations. First, all of them require linearity of the mean relationships among predictors along the central space; that is, $E(X|\beta^T X)$ is required to be linear in $\beta^T X$. When this condition fails, the methods may produce estimates that are $\sqrt{n}$-consistent for directions outside $\mathcal{S}_{Y|X}$. Since violations of the condition cannot be diagnosed prior to estimating $\beta$, it is common to pose the more restrictive assumption that $X$ be elliptically distributed. Ellipticity guarantees linearity of the mean relationships among predictors along any subspace, and can be at least partially diagnosed and remedied. In practice, one searches for curved patterns among predictors through scatter-plot matrices, and transforms predictors as to linearize these patterns e.g. through joint normalizing transformations (Cook, 1998). However, graphical detection of curvatures becomes cumbersome when the number of predictors is very high, and there are often limits to the extend to which curved patterns can be linearized by predictor transformations. Second, even when ellipticity holds, none of the global methods is guaranteed to be exhaustive: the estimates are $\sqrt{n}$-consistent for directions in $\mathcal{S}_{Y|X}$, but they may not recover the whole space, i.e., under ellipticity, the population-level subspace estimated by each of OLS, SIR, PHD or SAVE can be a proper subspace of $\mathcal{S}_{Y|X}$. This is arguably one of the most important shortcomings of these methods. An instance is the heavy reliance of methods such as OLS and SIR on linear trends in the dependence of $Y$ on $X$. For example, if $Y = (\beta^T X)^2 + \epsilon$ with $\beta \in \mathbb{R}^p$ and $X \sim N(0, I_p)$, OLS and SIR will estimate $0 \in \mathcal{S}_{Y|X} = Span(\beta)$, but fail to detect $\beta$ itself.

At the opposite side of the spectrum are adaptive methods that exploit *local* features of the dependence of $Y$ on $X$ (Xia, Tong, Li, and Zhu, 2002). The strength of these methods is that they require much weaker assumptions on the distribution $X$ (virtually none). However, because they employ multivariate kernels that shrink with the sample size, their convergence rates are generally slower than $\sqrt{n}$. In addition, these methods are computationally intensive, as they iterate between non-parametric estimation of a multivariate unknown function and numerical maximization of the estimated function over a potentially high-dimensional matrix of parameters.

In this paper, we postulate the following location structure,

$$Y = f(\beta^T X) + \epsilon \; , \;\; \epsilon \perp\!\!\!\perp X, E(\epsilon) = 0 \qquad (2)$$

where $f(\cdot)$ is an unknown function, and propose a novel approach for estimating $\beta$ based on estimating the contour directions of negligible variation for the response surface. These directions span the orthogonal complement of the central space, and can be extracted according to a measure of variation in the response, leading to what we call *General Contour Regression* (GCR). Unlike traditional global methods such as OLS, SIR, PHD and SAVE, GCR guarantees exhaustive estimation of the central space under ellipticity of $X$ and very mild additional assumptions. GCR also proves to be robust to violations of the ellipticity assumption. Moreover, unlike some of the local methods, GCR achieves $\sqrt{n}$-consistency and is computationally inexpensive. In fact, GCR does not involve iterative optimizations of complicated objective functions. Its working relies on a logic analogous to a one-dimensional kernel whose range can extend from very local to almost global, thus making effective use of the available sample size.

## 2 General Contour Regression

Our approach is based on the following simple observation: consider the regression function $E(Y|X) = f(\beta^T X)$, if $X_i$ and $X_j$ satisfy $\beta^T(X_i - X_j) = 0$, then $X_i$ and $X_j$ are on the same contour line of $f(\cdot)$. Therefore, we can use the orthogonal complement of the linear subspace spanned by such $X_i - X_j$ differences to try and recover $\beta$. Since $\beta$ is unknown, we cannot directly use $\beta^T(X_i - X_j) = 0$ as the criterion to identify relevant $X_i - X_j$ differences. One approach is to use

$$|Y_j - Y_i| = |(f(\beta^T X_i) + \epsilon_i) - (f(\beta^T X_j) + \epsilon_j)| \le c \quad (3)$$

where $c$ is a small constant (this is called simple contour regression in Li, Zha and Chiaromonte, 2003). If the regression function is non-monotone, this inequality may also pick up directions of sizeable response variation. Under ellipticity, these directions average out and thus do not introduce systematic biases. However, "wrong" directions do tend to decrease efficiency by blurring up "right" ones. We consider the following illustrative example.

**Example 2.1** Suppose $X = (X^{(1)}, X^{(1)})^T \sim N(0, I_2)$ and $Y = (X^{(2)})^2 + \epsilon$, with $\epsilon \perp\!\!\!\perp X$ and $\epsilon \sim N(0, \sigma^2)$. For this regression $\mathcal{S}_{Y|X}$ is the one-dimensional span of $\beta = (0, 1)$. We generated twenty observations $(X_i, Y_i)$ $i = 1, \ldots 20$ with $\sigma = 0.3$, and took $c$ in (3) to be 0.5. In the left panel of Figure 1, any two points $(X_i, X_j)^T \in \mathbb{R}^2$ satisfying $|Y_i - Y_j| \le 0.5$ were connected by a solid line segment. We see that, though most of the segments are horizontal (i.e. aligned with the true contour direction), there are many segments pointing to arbitrary directions. This is because $Y$ is roughly U-shaped and the inequality $|Y_i - Y_j| \le 0.5$ does not discriminate between the segments aligned with the contour and those across the U-shaped surface that also have small increments in $Y$. Though the arbitrary directions tend to average out due to the ellipticity of the distribution of $X$, they make the picture less sharp, and the method less efficient.

To overcome this drawback we replace the contour identifier $|Y_i - Y_j| \le c$ with a more sensitive one. Consider the variance of $Y$ along the line through $X_i$ and $X_j$. Formally, let $\ell(t; X_i, X_j) = (1-t)X_i + tX_j, t \in \mathbb{R}$ be the straight line that goes through $X_i$ and $X_j$, and

$$V(X_i, X_j) = \text{var}(Y|X = \ell(t; X_i, X_j) \text{ for some } t).$$

For a more concrete expression, let $\delta = \{\delta(X_i, X_j)\}$ be the $p \times (p-1)$ matrix $(\delta_1, \ldots \delta_{p-1})$ whose columns form a basis in $(X_j - X_i)^\perp$. Then, $V(X_i, X_j)$ can be re-expressed as

$$V(X_i, X_j) = \text{var}\left(Y|\delta^T(X_i, X_j)X = \delta^T(X_i, X_j)X_i\right). \quad (4)$$

We intend to identify contour directions requiring this conditional variance to be small, so our next task is to construct a sample estimate of $V(X_i, X_j)$. We will denote the line $\ell(\cdot; X_i, X_j)$ by $\ell(X_i, X_j)$. For any $X_k$, let $d(X_k, \ell(X_i, X_j))$ be the Euclidean distance between $X_k$ and the line $\ell(X_i, X_j)$; that is

$$d(X_k, \ell(X_i, X_j)) = \min_{t \in \mathbb{R}} \|X_k - \ell(t; X_i, X_j)\|,$$

where $\|\cdot\|$ stands for the Euclidean norm. It is easy to see that the above is the same as

$$\|X_k - X_i\|^2 - \frac{\left\{(X_k - X_i)^T(X_j - X_i)\right\}^2}{\|X_j - X_i\|^2}.$$

For any $\rho > 0$, we define the cylinder of radius $\rho$ connecting $X_i$ and $X_j$ to be

$$C_{ij}(\rho) = \{X_k : d(X_k, \ell(X_i, X_j)) < \rho, k = 1, \ldots n\}.$$

According to this definition, each cylinder contains at least 2 points in the sample. Next, we estimate the variance of $Y$ along such cylinders. Let $\#C_{ij}(\rho)$ be the number of points in the cylinder $C_{ij}(\rho)$, and let

$$\widehat{V}(X_i, X_j; \rho) = \frac{1}{\#C_{ij}(\rho)} \sum_{X_k \in C_{ij}(\rho)} (Y_k - \bar{Y}_{ij}(\rho))^2,$$

$$\bar{Y}_{ij}(\rho) = \frac{1}{\#C_{ij}(\rho)} \sum_{X_k \in C_{ij}(\rho)} Y_k.$$

We can now identify contour directions requiring $\widehat{V}(X_i, X_j; \rho)$ to be smaller than a certain threshold.

**Example 2.2** The right panel of Figure 1 contains the same sample points as the left panel, and shows line segments picked up by $\widehat{V}(X_i, X_j; \rho) \le c$, with $c = 0.5$ and $\rho = 0.3$. We can see that many of the segments pointing to random directions in the left panel have been removed. To provide a quantitative comparison, we calculated the first principal component of the vectors represented by line segments in each figure. For the left panel the component is $(0.9169, 0.3991)$; wheras for the right panel, it is $(0.9991, -0.0417)$. The latter is much closer to the direction $(1, 0)$, which is the population vector orthogonal to $\mathcal{S}_{Y|X}$. The simulation studies in Section 5 will confirm this improvement.

With the variance of $Y$ along the line through $X_i$ and $X_j$ thus estimated, we can now summarize our GCR algorithm for constructing the estimator of $\mathcal{S}_{Y|X}$.

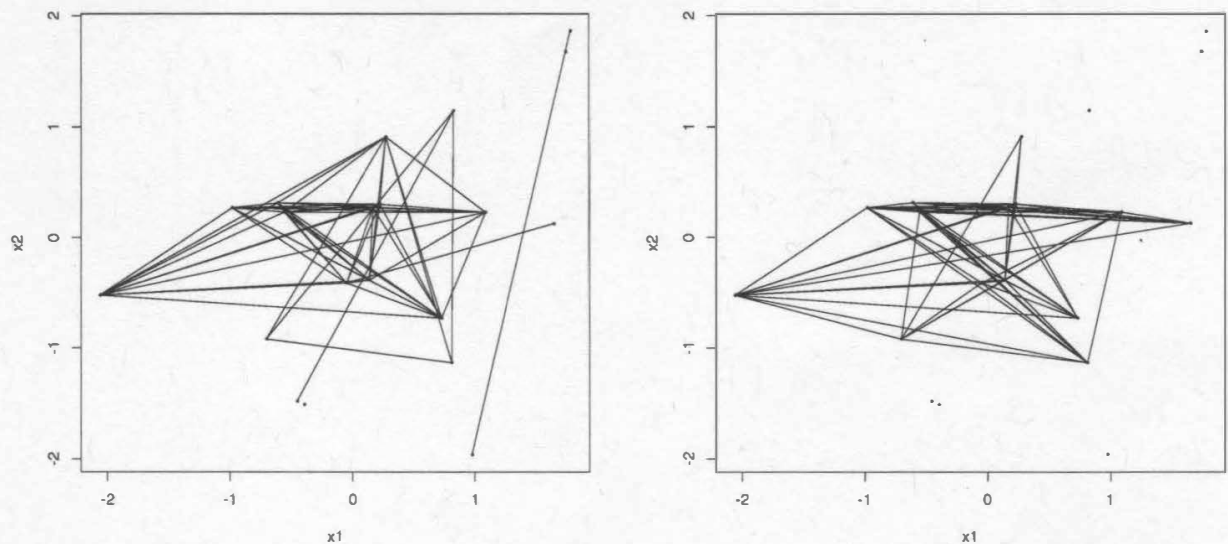

Figure 1: (left) Identified directions using $|Y_i - Y_j| \leq c$ and (right) Identified directions using $\widehat{V}(X_i, X_j; \rho) \leq c$.

- Compute sample mean and variance matrix of the predictor $X$

$$\hat{\mu} = n^{-1} \sum_{i=1}^{n} X_i, \quad \widehat{\Sigma} = n^{-1} \sum_{i=1}^{n} (X_i - \hat{\mu})(X_i - \hat{\mu})^T,$$

which will be used to standardize the predictor vectors.

- Compute the matrix-valued U-statistic

$$\widehat{F}(c) = \frac{1}{\binom{n}{2}} \sum_{(i,j) \in N} D_{ij} I(\widehat{V}(X_i, X_j; \rho) \leq c), \quad (5)$$

where $N$ is the index set $\{(i,j) : i = 2, \ldots n; j = 1, \ldots i-1\}$, $D_{ij} = (X_j - X_i)(X_j - X_i)^T$, and $I(\cdot)$ is the indicator function. $\widehat{F}(c)$ contains the sum of the out-products of those segments selected by the cylinder contour identifier.

- Compute the spectral decomposition of $\widehat{\Sigma}^{-1/2}\widehat{F}(c)\widehat{\Sigma}^{-1/2}$ and let $\hat{\gamma}_{p-q+1}, \ldots \hat{\gamma}_p$ be the eigenvectors corresponding to the smallest $q$ eigenvalues.

- The span of these eigenvectors estimates $\mathcal{S}_{Y|Z}$, where $Z$ is the standardized version of $X$. Thus, our estimate of the central space is

$$\widehat{\mathcal{S}}_{Y|X} = Span(\widehat{\Sigma}^{-1/2}\hat{\gamma}_{p-q+1}, \ldots \widehat{\Sigma}^{-1/2}\hat{\gamma}_p).$$

## 3 Population-level Exhaustiveness

Assume that $X$ is already standardized to $E(X) = 0$ and $var(X) = I_p$ (so $Z$ is $X$ itself). The population version of the matrix $\widehat{F}(c)$ in (5) is

$$F(c) = E[(X - \tilde{X})(X - \tilde{X})^T I(V(X, \tilde{X}) \leq c)]. \quad (6)$$

We will demonstrate that, for sufficiently small $c$, the eigenvectors corresponding to the $q$ smallest eigenvalues of $G(c)$ span $\mathcal{S}_{Y|X}$. Here, $(\tilde{X}, \tilde{Y})$ indicates an independent copy of $(X, Y)$. Detailed proofs of the statements in this section are omitted, and can be found in Li, Zha and Chiaromonte (2003). First, we need the following assumption.

**Assumption 3.1** *For any choice of vectors $v \in \mathcal{S}_{Y|X}$ and $w \in (\mathcal{S}_{Y|X})^\perp$ such that $\|v\| = \|w\| = 1$, and any sufficiently small $c > 0$, we have*

$$\begin{aligned} var\left[w^T(\tilde{X} - X) \middle| V(X, \tilde{X}) \leq c\right] &> \\ var\left[v^T(\tilde{X} - X) \middle| V(X, \tilde{X}) \leq c\right]. \end{aligned} \quad (7)$$

This assumption is a reasonable one: because the conditional distribution of $Y|X$ depends on $v^T X$ but not on $w^T X$, we expect $Y$ to vary more with $v^T X$ than it does with $w^T X$. Hence, intuitively, within the same increment of $Y$, $w^T X$ should vary more than $v^T X$ does. We now deduce population exhaustiveness under this assumption, and we do so for a spherical predictor without loss of generality.

**Theorem 3.1** *Suppose that $X$ has an elliptical distribution with $E(X) = 0$ and $var(X) = I_p$. Then, under Assumption 3.1, for sufficiently small $c > 0$, the eigenvectors of $F(c)$ corresponding to its $q$ smallest eigenvalues span the central space $\mathcal{S}_{Y|X}$.*

In reference to the location structure in (2), we derive a sufficient condition for Assumption 3.1.

Note that, since $\mathcal{S}_{Y|X} = \text{span}(\beta)$, for a $p \times r$ matrix $\delta$ ($r \leq p$) we will have

$$var(f(\beta^T X)|\delta^T X) > 0 \quad (8)$$

unless span($\beta$) $\subset$ span($\delta$); that is, $f(\beta^T X)$ is not a function of $\delta^T X$ unless $\delta$ spans a space containing the central subspace.

**Theorem 3.2** *Suppose that $X$ has an elliptically-contoured distribution with $E(X) = 0$ and $var(X) = I_p$. Then Assumption 3.1 is satisfied for all sufficiently small $c > 0$ for which $\{(x, \tilde{x}) : V(x, \tilde{x}) \leq c\}$ is a non-empty set.*

It is also interesting to make a comparison between contour regression and SAVE. Consider again the case of $Y = X_1^2 + \epsilon$, with $X \in \mathbb{R}^2$ uniformly distributed on a unit disc centered at the origin. Because the response is U-shaped in $X_1$, each slice around $Y = y$ will identify data points forming two parallel strips aligned with the $X_2$ axis. Roughly speaking, SAVE performs a Principal Component Analysis on all the points in a slice, after centering them at the origin. Thus, if the strips are long and close, as is the case for small $y$ values, then the principal component is aligned with the $x_2$ axis as desired. However, if the two strips are short and far apart, as is the case for large $y$ values, then the principal component is aligned with the $x_1$ axis, yielding the wrong direction. GCR avoids this problem by taking into account what happens "across slices": line segments connecting two points across the U-shape will be excluded from the estimation because the response variance along the line through the two points is large.

## 4 Robustness against non-ellipticity

The population exhaustiveness of our contour-based methodology relies on ellipticity of the predictor distribution. This is because in the theoretical development we have treated the constant $c$ in (5) as fixed with respect to the sample size $n$. Ellipticity of the distribution of $X$ helps to balance out the effect of those line segments not aligned with the contour directions. However, especially when using GCR, whose contour identifier is more sensitive, we can obtain good performance even under violations of ellipticity.

Here, we motivate this robustness from a theoretical viewpoint. We will show that, postulating as always the location structure in (2), the eigenvectors corresponding to the smallest $p - q$ eigenvalues of the matrix

$$A = E\left((\tilde{X} - X)(\tilde{X} - X)^T \Big| V(X, \tilde{X}) = \sigma^2\right)$$

span the orthogonal complement of the central subspace, $(\mathcal{S}_{Y|X})^\perp$, even when $X$ is not elliptical. This suggests that if we let $c$ decrease to $\sigma^2$ as $n$ increases, then the eigenvectors corresponding to the smallest $p-q$ eigenvalues of $\widehat{F}(c)$ in (5) (after appropriate transformation by $\widehat{\Sigma}^{-1/2}$) will tend to recover the whole $\mathcal{S}_{Y|X}$, regardless of the shape of the distribution of $X$. In practice, if we make $c$ small (i.e. close to the smallest value of $\widehat{V}(\widehat{X}_i, \widehat{X}_j)$ in (5)), then GCR is likely to estimate the central subspace exhaustively and effectively even if the shape of $X$ does not help the process by averaging out erroneous directions, as is the case under ellipticity.

**Theorem 4.1** *Suppose that $X$ is a continuous random vector with an open support $\mathcal{X} \subset \mathbb{R}^p$. Then the matrix $A$ has exactly $p - q$ zero eigenvalues, and their corresponding eigenvectors span $(\mathcal{S}_{Y|X})^\perp$. In symbols,*

$$\ker(A) = \mathcal{S}_{Y|X}$$

*where $\ker(A) = \{h \in \mathbb{R}^p : Ah = 0\}$ is the kernel of $A$.*

PROOF. Note that $(\tilde{X} - X)$ is orthogonal to span($\beta$) = $\mathcal{S}_{Y|X}$ if and only if span($\beta$) $\subset$ span($\delta(X, \tilde{X})$), which happens if and only if $V(X, \tilde{X}) = \sigma^2$. Thus, conditioning on $V(X, \tilde{X}) = \sigma^2$, $(\tilde{X} - X)$ is orthogonal to span($\beta$). It follows that, whenever $h$ belongs to span($\beta$), $Ah = 0$, and thus span($\beta$) $\subset$ ker($A$).

Conversely, suppose $h$ belongs to ker($A$). Then

$$h^T A h = E[(h^T(\tilde{X} - X))^2 | V(X, \tilde{X}) = \sigma^2] = 0.$$

Thus, whenever $V(X, \tilde{X}) = \sigma^2$, $h$ is orthogonal to $(\tilde{X} - X)$. Equivalently, whenever $(\tilde{X} - X)$ is orthogonal to span($\beta$), $(\tilde{X} - X)$ is orthogonal to $h$. In other words, if we let $\mathcal{X}^* = \{\tilde{x} - x : \tilde{x} \in \mathcal{X}, x \in \mathcal{X}\}$, then

$$\mathcal{X}^* \cap (\text{span}(\beta))^\perp \subset \mathcal{X}^* \cap (\text{span}(h))^\perp.$$

However, because $\mathcal{X}$ is an open set, $\mathcal{X}^*$ is an open set containing 0. It can be shown that $(\text{span}(\beta))^\perp \subset (\text{span}(h))^\perp$, or equivalently $h \in \text{span}(\beta)$, as desired. □

## 5 Experimental Results

We now compare the performance of GCR with that of well known existing dimension reduction methods that have $\sqrt{n}$-consistency, such as OLS, SIR, PHD, and SAVE. We measure the distance between two subspaces $\mathcal{S}_1$ and $\mathcal{S}_2$ using

$$\text{dist}(\mathcal{S}_1, \mathcal{S}_2) = \|P_{\mathcal{S}_1} - P_{\mathcal{S}_2}\|,$$

where $P_{\mathcal{S}_i}$ is the orthogonal projection onto $\mathcal{S}_i$, $i = 1, 2$, and $\|\cdot\|$ is the spectral norm.

In the following examples, $X$ is a standard multivariate normal random vector and the dimension of the central subspace is taken to be $q = 2$. The sample size $n$ is 100 in each example, and performance statistics are based on 500 runs. We need to determine the number of predictor differences to include in the Principal Component Analysis. We choose to use the $2qn$ predictor differences whose corresponding response differences are smallest in absolute value. We also need to choose the tube size $\rho$ for computing $\widehat{V}(X_i, X_j; \rho)$. In all examples the tube size is taken to be $\rho = 0.01$. We have not optimized the choice of these two parameters; further research is needed in this regard.

Table 1: Comparison of GCR and other estimates for Example 5.1

|  | GCR | | SIR | | SAVE | | PHD | |
|---|---|---|---|---|---|---|---|---|
| $\sigma$ | DIST | SE | DIST | SE | DIST | SE | DIST | SE |
| 0.1 | 0.16 | 0.07 | 0.78 | 0.24 | 0.43 | 0.25 | 0.80 | 0.21 |
| 0.4 | 0.20 | 0.08 | 0.79 | 0.23 | 0.54 | 0.27 | 0.79 | 0.21 |
| 0.8 | 0.32 | 0.16 | 0.80 | 0.23 | 0.73 | 0.25 | 0.79 | 0.21 |

**Example 5.1** Consider the regression $Y = X_1^2 + X_2 + \sigma\epsilon$, where $X$ is a 4-dimensional standard multivariate normal vector, $\epsilon$ is a standard normal random variable independent of $X$, and $\sigma$ is taken to be 0.1, 0.4, and 0.8. Here, the central subspace is of dimension $q = 2$, and is spanned by the vectors $(1, 0, 0, 0)$ and $(0, 1, 0, 0)$. The response surface is quadratic in the direction of the former, and linear in that of the latter. We compare GCR with SIR, SAVE, and PHD. A common sample is used for all four estimators in each simulation. Simulation results are summarized in Table 1, with distance averages and standard errors presented, respectively, in the DIST and SE columns.

Table 1 indicates that GCR performs better than SIR, SAVE, and PHD. Intuitively, this is because SIR does not perform well when there is no linear trend, and thus fails to pick up the first direction $(1, 0, 0, 0)$, whereas PHD, and to a lesser extent SAVE, do not perform well when there is no quadratic trend, and thus fail to accurately estimate of the second direction $(0, 1, 0, 0)$. In contrast, GCR provides comprehensive estimates of the central subspace, confirming the theoretical results discussed above. Note that SAVE performs better than SIR and PHD — inspection of a few typical cases (results not shown) suggest that SAVE deals with linear trends better than PHD. Nevertheless, it remains much less accurate than GCR.

In the next example, we compare methods using a more complex regression surface than that in Example 5.1. In this more complex situation, SIR, SAVE and PHD may also effectively pick up both directions.

**Example 5.2** Consider the regression

$$Y = X_1/(0.5 + (X_2 + 1.5)^2) + (1 + X_2)^2 + \sigma\epsilon,$$

where $X$, $\epsilon$, and $\sigma$ are as defined in Example 5.1. Again, $q = 2$ and the central subspace is spanned by the vectors $(1, 0, 0, 0)$ and $(0, 1, 0, 0)$ – all specifications other than the regression surface are the same as in Example 5.1. Simulation results are summarized in Table 2, and show that notwithstanding the increased complexity of the regression, GCR still provides a substantial improvement over SIR, SAVE and PHD.

In section 3 we stated that GCR is robust against departures from ellipticity in the distribution of $X$. We now compare GCR with OLS, PHD, SIR, and SAVE when the distribution of $X$ is not elliptical.

**Example 5.3** Consider the nonlinear regression $Y = \sin[(\pi X_2 + 1)^2] + \sigma\epsilon$, with predictor $X$ in $\mathbb{R}^4$. The central subspace has dimension $q = 1$ and is spanned by the vector $(0, 2, 0, 0)$. We take $X$ to be uniformly distributed on the following set:

$$[0, 1]^4 \setminus \{x \in \mathbb{R}^4 : x_i \leq \tau, \ i = 1, 2, 3, 4, \quad \tau = 0.7\}.$$

This is a 4-dimensional cube with a corner removed. As in the previous examples, $\epsilon$ is standard normal independent of $X$, and the standard deviation $\sigma$ is fixed at 0.1, 0.2, and 0.3. Simulation results are summarized in Table 3.

We see that GCR performs better than the other estimators. Also, among the latter SIR and SAVE appear to be more robust than OLS and PHD against departures from ellipticity of $X$.

**Example 5.4** We consider data collected for Massachusetts four-year colleges in 1995, in an attempt to investigate how the percentage of freshmen that graduate (Grad) depends on variables measuring quality of incoming students and features of the colleges. The data is provided as an example data set in MINITAB (release 13.32, data directory STUDNT12). We restricted attention to $n = 46$ colleges and $p = 7$ predictors, which are: the percentage of freshmen that were among the top 25% percent in their graduating high school class (Top25), the median mathematics SAT score (MSAT), the median verbal SAT score (VSAT), the percentage of applicants accepted by the college (Accept), the percentage of accepted applicants who enroll (Enroll), the student-to-faculty ratio (SFRatio), and the out-of-state tuition (Tuition).

Table 2: Comparison of GCR and other estimates for Example 2

|   | GCR | | SIR | | SAVE | | PHD | |
|---|------|------|------|------|------|------|------|------|
| $\sigma$ | DIST | SE | DIST | SE | DIST | SE | DIST | SE |
| 0.1 | 0.28 | 0.15 | 0.39 | 0.21 | 0.61 | 0.26 | 0.71 | 0.25 |
| 0.4 | 0.33 | 0.18 | 0.40 | 0.21 | 0.65 | 0.26 | 0.70 | 0.25 |
| 0.8 | 0.45 | 0.25 | 0.49 | 0.24 | 0.73 | 0.24 | 0.73 | 0.24 |

Table 3: Comparison when $X$ has non-elliptical distribution

|   | GCR | | OLS | | SIR | | SAVE | | PHD | |
|---|------|------|------|------|------|------|------|------|------|------|
| $\sigma$ | DIST | SE | DIST | SE | DIST | SE | DIST | SE | DIST | SE |
| 0.1 | 0.10 | 0.05 | 0.17 | 0.07 | 0.24 | 0.10 | 0.13 | 0.06 | 0.14 | 0.08 |
| 0.2 | 0.12 | 0.06 | 0.19 | 0.09 | 0.29 | 0.12 | 0.18 | 0.08 | 0.22 | 0.12 |
| 0.3 | 0.20 | 0.14 | 0.22 | 0.10 | 0.36 | 0.16 | 0.22 | 0.10 | 0.34 | 0.20 |

A scatter-plot matrix of the data (not shown) reveals obvious curvatures in the mean dependencies among predictors. This violation of ellipticity, while in principle troublesome for both traditional methods and contour-based methods, is likely to be better withstood by the latter, as discussed in Section 4. Moreover, marginal regression plots (response vs individual predictors) suggest the existence of non-trivial patterns in the dependence of Grad on the predictors. As discussed in Section 1, if these patterns lack a marked linear component along some of the directions they comprise, these directions may be missed by non-exhaustive methods that rely heavily on linear trends (e.g. SIR) even when ellipticity holds.

We apply GCR to the data set, taking the tube size to be $\rho = 0.03$ and including $4n = 184$ pairs of predictor differences with the smallest $\widehat{V}(\widehat{Z}_i, \widehat{Z}_j; \rho)$ values. This gives eigenvalues (from smallest to largest) 2.1866, 3.6160, 7.6274, 7.7670, 8.6623, 9.6466 and 10.5777. Even though we have not yet developed a rigorous theory for dimensional inference with GCR, the clear separation between the first two eigenvalues and the following five allows us to infer the existence of two relevant directions, which correspond to the estimated linear combinations

$$\text{GCR1} = -0.6331(\text{Top25}) + 0.0168(\text{MSAT})$$
$$+0.1519(\text{VSAT}) + 0.4068(\text{Accept})$$
$$-0.0726(\text{Enroll}) + 0.6365(\text{SFRatio})$$
$$+0.0004(\text{Tuition})$$

$$\text{GCR2} = +0.1915(\text{Top25}) - 0.0605(\text{MSAT})$$
$$+0.1336(\text{VSAT}) + 0.8642(\text{Accept})$$
$$-0.1622(\text{Enroll}) - 0.4106(\text{SFRatio})$$
$$+0.0011(\text{Tuition})$$

Views of the 3D plot of Grad vs GCR1 and GCR2 are given in Figure 2, revealing a peculiar "coiled" structure for the dependence of the response on the reduced predictors. While the linear component along GCR1 is strong (R-square approximately 56%), that along GCR2, which shows the bending of the coil, is much weaker (R-square approximately 8%). Indeed, SIR applied to the same data solidly detects the first direction, while producing ambiguous results on the existence of a second relevant direction. In SIR, sequential asymptotic chi-square testing can be employed for dimensional inference under ellipticity of $X$ (see Li, 1991 and Cook, 1998, for details). When applied to these data the tests produce p-values below 0.01 for SIR1 (first vector from SIR), regardless of the number of slices employed in the SIR algorithm. However, p-values for SIR2 (second vector from SIR) range between 0.10 and 0.30 depending on the number of slices.

## 6 Conclusions

The main strength of the contour-based method introduced in this paper is that, under very mild conditions, it achieves exhaustive estimation of the central subspace at the $\sqrt{n}$-convergence rate. In comparison with existing global estimators such as OLS, PHD, SIR, and SAVE, contour-based estimators are more comprehensive, capable of picking up all directions in the central subspace without relying on special response patterns (e.g. linear or quadratic trends). At the same time, unlike adaptive estimators, contour-based estimators guarantee $\sqrt{n}$-convergence regardless of the original dimension $p$ and the structural dimension $q$. Furthermore, they are computationally simple – the level of computational burden being essentially that of Principal Component Analysis. In this respect they are similar to global methods. In particular, they do not require iterative maximization of a multivariate nonparametric function. This can be a substantial advantage, particularly if the dimension is large, or if multiple local maxima are present in the iterative maximization. Finally, while the theoretical developments we presented do assume an elliptical distribution for $X$, contour-based estimators retain a degree of robustness if ellipticity is violated.

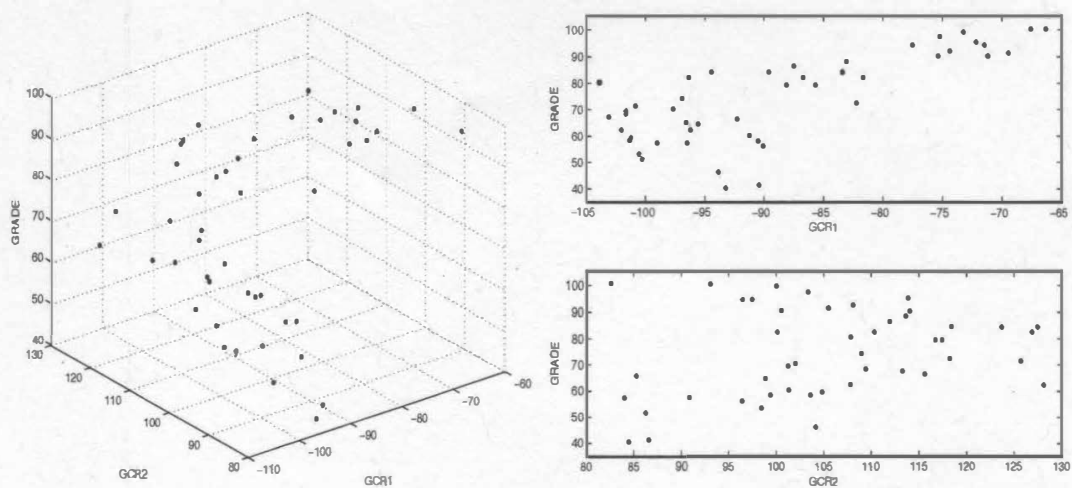

Figure 2: Perspective plot of Grad against CR1 and CR2 (left) and Scatter plots of Grad against CR1 (upper) and CR2 (lower), respectively (right).

This of course does not imply that contour-based estimators will outperform other estimators under all circumstances. For example, OLS is the maximum likelihood estimator if the regression surface is linear, and tends to perform very well if the surface is nearly linear or clearly monotone. Similarly favorable circumstances exist for PHD, SIR, and SAVE as well.

The basic ideas in contour regression raise many questions that have not been addressed within the the scope of this paper. In particular, the asymptotic properties of GCR, as well as test statistics for determining the structural dimension $q$, have not yet been developed. We do expect that $\sqrt{n}$-rate can be achieved by GCR if the threshold $c$ is taken as fixed, because this in effect includes in the computation a number of line segments proportional to the total number of observation pairs. Other useful developments will concern the asymptotic behavior of GCR when the threshold $c$ is allowed to go to zero as the sample size $n$ tends to infinity. Theorem 4.1 suggests that the correct asymptotic behavior would still be guaranteed, without assuming ellipticity of $X$. However, in this case we do not expect a $\sqrt{n}$ convergence rate – at least not for all dimensions. Finally, to further improve efficiency it may be helpful to experiment with windows other than the current rectangular ones in selecting contour vectors. We leave these issues to future studies.